%% file: main.tex
\documentclass[10pt,twocolumn,letterpaper]{article}

\usepackage{wacv}              


\definecolor{wacvblue}{rgb}{0.21,0.49,0.74}
\usepackage[pagebackref,breaklinks,colorlinks,allcolors=wacvblue]{hyperref}
\usepackage{algorithm}
\usepackage{algpseudocode}
\usepackage{multirow}
\usepackage[table]{xcolor}

\def\wacvPaperID{2402} 
\def\confName{WACV}
\def\confYear{2026}

\title{Quality-Driven and Diversity-Aware Sample Expansion for \\ Robust Marine Obstacle Segmentation}

\author{Miaohua Zhang\textsuperscript{1}, 
Mohammad Ali Armin\textsuperscript{1}, 
Xuesong Li\textsuperscript{2}, 
Sisi Liang\textsuperscript{1}, \\
Lars Petersson\textsuperscript{1}, 
Changming Sun\textsuperscript{1}, 
David Ahmedt-Aristizabal\textsuperscript{1}, 
Zeeshan Hayder\textsuperscript{1}\\
CSIRO Data61\textsuperscript{1}, 
CSIRO Agriculture \& Food\textsuperscript{2} \\
{\tt\small \{Miaohua.Zhang; Ali.Armin; David.Ahmedtaristizabal; Zeeshan.Hayder\}@data61.csiro.au}
}

\begin{document}
\maketitle
\input{sec/0_abstract}    
\input{sec/1_intro}

\input{sec/2_formatting}
\input{sec/3_finalcopy}

\clearpage
{
    \small
    \bibliographystyle{ieeenat_fullname}
    \bibliography{main}
}

\end{document}

%% file: sec/0_abstract.tex
\begin{abstract}
Marine obstacle detection demands robust segmentation under challenging conditions, such as sun glitter, fog, and rapidly changing wave patterns. These factors degrade image quality, while the scarcity and structural repetition of marine datasets limit the diversity of available training data. Although mask-conditioned diffusion models can synthesize layout‑aligned samples, they often produce low-diversity outputs when conditioned on low-entropy masks and prompts, limiting their utility for improving robustness.
In this paper, we propose a quality-driven and diversity-aware sample expansion pipeline that generates training data entirely at inference time, without retraining the diffusion model. 
The framework combines two key components: 
(i) a class-aware style bank that constructs high-entropy, semantically grounded prompts, 
and (ii) an adaptive annealing sampler that perturbs early conditioning, while a COD‑guided proportional controller regulates this perturbation to boost diversity without compromising layout fidelity. Across marine obstacle benchmarks, augmenting training data with these controlled synthetic samples consistently improves segmentation performance across multiple backbones and increases visual variation in rare and texture-sensitive classes. 
\end{abstract}

%% file: sec/1_intro.tex
\vspace{-0.5cm}
\section{Introduction}
\label{sec:intro}

Autonomous surface vessels (USVs) must reliably segment obstacles in marine scenes dominated by large expanses of water, low-contrast horizons, and open sky. These environments present fast-changing visual conditions, where wave geometry, foam patterns, fog, haze, and sun glitter can distort critical visual cues~\cite{zust2022temporal,ahmed2023vision}. Accurate segmentation becomes even more challenging when lighting conditions vary rapidly and when small obstacles appear infrequently~\cite{vzust2023lars}.
\begin{figure}[t]
    \centering
    \includegraphics[width=\linewidth]{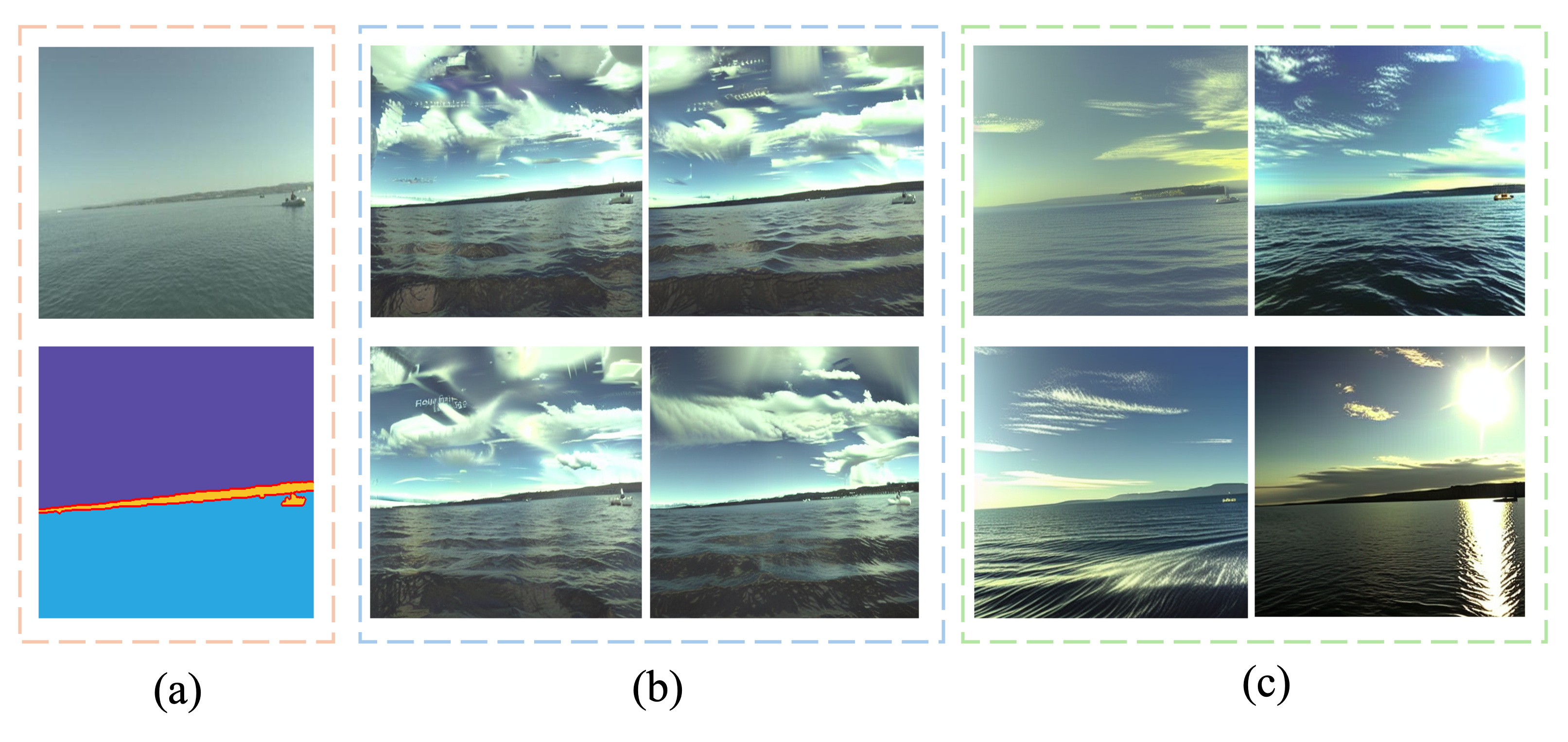} 
    \vspace{-0.8cm}
    \caption{(a) Original image and its mask; 
    (b) generated images using a simple prompt; 
    (c) generated images using the proposed style bank and Adaptive Annealed  Sampling (AAS) strategy.}
    \label{fig:concept}
    \vspace{-0.5cm}
\end{figure}

Public marine datasets reflect these challenges but remain limited in size and diversity, with highly structured scene layouts. For example, MaSTr1325~\cite{bovcon2019mastr1325} provides 1,325 pixel-wise labeled images collected over two years by a coastal USV. While it captures diverse marine conditions, most scenes follow a repetitive structure: a broad water region, a narrow horizon line, and an overlying sky band. Obstacles are typically small and imbalanced in distribution~\cite{liu2021estimating}. Similarly, the MODD/MODD2 datasets~\cite{bovcon2021mods} increase the difficulty by incorporating multi-modal sensor recordings, adverse weather, and strong specular reflections. These datasets focus their annotations on water-edge localization and obstacle detection near the boat’s trajectory, which are crucial for safe navigation.

Despite these efforts, segmentation models trained on these datasets still underperform. Existing benchmarks continue to highlight failures in the presence of various patterns ~\cite{vzust2022temporal}. Such effects significantly degrade detection results, often leading to false positives and missed detections~\cite{yalccin2024impact}. 
While expanding the training data is a standard strategy for improving model generalization, in the maritime domain collecting and labeling diverse, high-resolution scenes remains prohibitively expensive and challenging~\cite{liu2021real}. 
At-sea operations are constrained by narrow weather windows as well as limited vessel availability,  making large-scale data acquisition difficult. Furthermore, annotating marine images at the pixel level is particularly costly: for reference, Cityscapes reports more than 1.5 hours per finely annotated image, and this burden becomes even heavier in marine environments due to imbalanced targets and frequent occlusions from waves, fog, or reflections~\cite{pan2024obstacle}.

To reduce annotation costs, some approaches employ weak labels, such as water-edge boundaries or coarse obstacle bounding boxes. While this reduces labeling time and can support weakly supervised segmentation, it does not fully capture the semantic complexity of small or partially occluded obstacles~\cite{vzust2022learning}. Prior work shows that even with strong priors, weakly supervised models remain vulnerable to failure under extreme appearance shifts, such as reflections, glitter, or low-light conditions~\cite{chen2025weakly,vzust2022temporal}.

Generative augmentation has therefore emerged as a promising alternative because of its powerful ability to improve visual diversity and sample quality while preserving geometric structure. Diffusion models have recently enabled controllable image synthesis for tasks such as image-to-image translation (e.g., Palette)~\cite{saharia2022palette}, semantic editing (SDEdit)~\cite{meng2021sdedit}, and structure-conditioned generation (ControlNet, T2I-Adapter)~\cite{zhang2023adding,mou2024t2i}. By conditioning on masks, edges, or depth maps, these models can generate realistic images that preserve spatial layouts, making them particularly suitable for segmentation pipelines. Indeed, mask-conditioned diffusion has improved downstream segmentation in aerial imagery~\cite{mcwilliams2024diffusion} and medical imaging~\cite{wu2024medsegdiff} by generating paired image-mask samples with controlled variations. However, such methods have not yet been effectively adapted to marine environments, where both appearance diversity (e.g., weather, lighting, texture) and sample fidelity (e.g., water, sky, obstacle layout) are essential.

Our work is motivated by the need to enhance visual quality and diversity of synthetic training data without sacrificing semantic consistency, particularly in domains with a limited set of semantic labels and imbalanced training data~\cite{ho2022classifier,bovcon2019mastr1325,vzust2023lars}. In such cases, semantic masks encode repetitive structures, and natural language prompts offer minimal variation, causing conditional diffusion models to generate samples with limited variability. This restricts the effectiveness of generative augmentation for downstream tasks such as marine obstacle detection, where changes in weather, lighting, and water appearance are critical.

To address these challenges, we propose a two-stage framework that separates training and inference responsibilities. Our method operates entirely at inference time and requires no retraining of the diffusion model. During training, the diffusion model is conditioned with minimal class-list prompts to maintain semantic clarity. At inference, we inject quality-aware appearance variation through a class-aware style prompting mechanism and an Adaptive Annealed Sampling (AAS) strategy. The style prompting mechanism leverages a domain-adaptive style bank that dynamically samples class-specific descriptors (e.g., ``stormy coastal sky'', ``choppy harbor waves''), enabling high-entropy yet semantically grounded prompts. 
In parallel, AAS adaptively modulates conditioning strength during early diffusion steps, guided by conditional-output discrepancy, to encourage exploration before converging on mask-aligned outputs. This modular design requires no retraining, integrates seamlessly with existing conditional diffusion backbones, and consistently yields visually diverse and high-quality samples that improve segmentation performance when used for data augmentation. As shown in Figure \ref{fig:concept}, our method significantly increases visual diversity while maintaining mask alignment, compared to generation with simple prompts.

The key contributions can be summarized as follows:
\begin{itemize}
  \item We introduce a plug-and-play framework that enhances sample quality and visual diversity in conditional image generation through structured prompting and adaptive sampling without requiring retraining of the diffusion model. This is especially valuable in domains like marine obstacle detection, where scene structure is repetitive and the label space is limited.
 \item We design a modular and extensible prompting strategy that composes prompts dynamically using a class-aware style bank. This enables semantically grounded yet visually diverse image synthesis and allows the method to generalize to datasets with different scene characteristics.
 \item We propose a novel sampling strategy that perturbs the conditioning signal during early diffusion steps, guided by a feedback controller that tracks the conditional-output discrepancy (COD). AAS injects targeted stochasticity to broaden appearance variation while preserving alignment with the input mask.
 \end{itemize}

%% file: sec/2_formatting.tex
\section{Related Works}
\label{sec:formatting}

\noindent\textbf{Traditional Augmentation for Segmentation.}
Traditional augmentation techniques, such as flipping, rotation, cropping, brightness/contrast jitter, and elastic deformations, remain strong baselines for semantic segmentation~\cite{wang2024comprehensive}. In medical and remote sensing tasks, aggressive deformations~\cite{xu2000image} and scale jitter are widely used to address data scarcity, while models like DeepLab~\cite{chen2017deeplab} routinely apply multiscale cropping and random resizing. Beyond manual transformations, learning-based methods explore richer transformation spaces. AutoAugment~\cite{cubuk2019autoaugment}, RandAugment~\cite{cubuk2020randaugment}, and TrivialAugment~\cite{muller2021trivialaugment} automate policy discovery; AugMix~\cite{hendrycks2019augmix} improves robustness to distribution shift; and Cutout~\cite{devries2017improved}, MixUp~\cite{zhang2017mixup}, and CutMix~\cite{yun2019cutmix} introduce sample mixing or occlusion priors that can help dense prediction when they are mask-aware.

While these methods are efficient and label-preserving, they reuse existing appearance and can violate scene realism, e.g., disrupting horizon lines or creating implausible water-sky boundaries. As a result, they struggle with long-tailed appearance variations like glitter, fog, or specular reflections, which frequently cause false positives in maritime segmentation. These limitations motivate the use of generative methods that preserve geometry while introducing new, realistic visual diversity.

\noindent\textbf{Generative Augmentation.}
Generative augmentation synthesizes new training pairs that diversify appearance while preserving label structure. Recent work shows that the effectiveness of synthetic data depends not just on realism, but also on visual diversity, semantic fidelity, and alignment with structural priors, especially in safety-critical domains like marine obstacle segmentation. 

Diffusion models are increasingly preferred for their controllability and visual quality. 
Palette~\cite{saharia2022palette} unified image-to-image tasks with conditional diffusion and outperformed GANs. SDEdit~\cite{meng2021sdedit} demonstrated how noise and re-denoising enable structure-preserving edits.. For spatial control, ControlNet~\cite{zhang2023adding} and T2I-Adapter~\cite{mou2024t2i} integrate masks, edges, or depth into large text-to-image models to ensure better layout alignment, which is essential for segmentation tasks.

Evidence from remote sensing~\cite{toker2024satsynth, sousa2025data} and medical imaging~\cite{wu2024medsegdiff} confirms that diffusion-based augmentation improves robustness, especially when generating anatomically or structurally consistent pairs.
These methods generate layout-preserving yet visually diverse samples. However, marine obstacle detection remains underexplored, despite its low-entropy layouts and extreme appearance variations (e.g., fog, glare, water patterns), which challenge existing generative models.

\noindent\textbf{Sampling and Diversity in Diffusion Models.}
A key challenge in conditional diffusion is balancing semantic fidelity and sample diversity. Classifier-free guidance (CFG)~\cite{ho2022classifier} enables a tunable trade-off: stronger guidance improves alignment but reduces diversity, especially problematic under low-entropy conditions like repetitive masks and generic prompts. Studies have shown that sampling and training choices affect output quality and mode coverage. Fast ODE samplers~\cite{lu2022dpm} improve inference efficiency without retraining, making them well-suited for large backbones. To enhance diversity, condition-annealed diffusion (CADS)~\cite{sadat2023cads} gradually weakens conditioning early and restores it later to preserve fidelity. Plug-and-Play diffusion~\cite{tumanyan2023plug} manipulates internal features to control layout or style without modifying the backbone. These works demonstrate that inference-time sampling strategies can broaden visual modes while preserving structure, which is critical for effective augmentation in domains like marine scenes, where conditioning inputs are low in entropy.

%% file: sec/3_finalcopy.tex
\section{Method}
\subsection{Problem Formulation}

Our goal is to generate visually diverse and perceptually realistic images that remain semantically consistent, conditioned on structured inputs, such as segmentation masks, and text prompts. This problem arises in marine obstacle detection, where semantic labels are limited and textual prompts often offer low diversity.

Let $\mathbf{M} \in \mathcal{M}$ denote the structured condition input (e.g., a segmentation mask), and $\mathbf{t} \in \mathcal{T}$ be a textual prompt describing the semantic content of the image. The objective is to learn a generative process that samples images $\mathbf{x} \in \mathcal{X}$ from the conditional distribution:
\begin{equation}
    p(\mathbf{x} \mid \mathbf{M}, \mathbf{t})
\end{equation}

In the context of diffusion models, generation is performed via a reverse denoising process starting from a Gaussian noise vector $\mathbf{x}_T \sim \mathcal{N}(0, I)$, and iteratively refining it through a learned denoising network over $T$ timesteps. The forward diffusion process defines a noising schedule:
\begin{equation}
    q(\mathbf{x}_t \mid \mathbf{x}_{t-1}) = \mathcal{N}(\mathbf{x}_t; \sqrt{1 - \beta_t} \, \mathbf{x}_{t-1}, \beta_t \, \mathbf{I})
\end{equation}
and the reverse process is learned via a neural network $\epsilon_\theta(\mathbf{x}_t, t, \mathbf{M}, \mathbf{t})$ that estimates the noise at each step. The training objective minimizes the expected denoising error across randomly sampled timesteps:
\begin{equation}
\scalebox{0.85}{$
    \mathcal{L}_{\text{DM}} = \mathbb{E}_{\mathbf{x}_0, \mathbf{M}, \mathbf{t}, \epsilon \sim \mathcal{N}(0, I), t \sim \mathcal{U}[1, T]} \left[ \left\| \epsilon - \epsilon_\theta(\mathbf{x}_t, t, \mathbf{M}, \mathbf{t}) \right\|_2^2 \right].
$}
\end{equation}
Here, $x_0$ denotes the original clean image, $x_t$ represents the noised version of $x_0$ obtained by applying a forward noise schedule at a randomly sampled timestep $t \in [1, T]$. The variable $\epsilon \sim \mathcal{N}(0, I)$ is the Gaussian noise added to produce $x_t$. The conditioning input $y$ typically corresponds to some guidance such as a segmentation mask or text prompt, and $\tau$ denotes auxiliary information, such as a style embedding. The function $\epsilon_\theta(x_t, t, y, \tau)$ is a neural network (usually a U-Net) that attempts to predict the added noise $\epsilon$, conditioned on the noisy input $x_t$, the timestep $t$, and the conditioning inputs $y$ and $\tau$.

However, existing conditional diffusion models often suffer from a lack of diversity during inference, particularly when using fixed conditioning inputs $\mathbf{M}$ and simple prompts $\mathbf{t}$. This issue is especially pronounced in real-world settings such as marine obstacle detection, where the available class information remains nearly consistent or too generic across different samples (e.g., sky, water, obstacle). When paired with simple prompts like ``this image contains sky, water, and obstacle'', the text-to-image model receives minimal high-frequency or stylistic guidance, leading to mode collapse in which generated images converge to visually similar patterns, such as repeated sky gradients, homogeneous wave textures, or similar terrain layouts. Such redundancy severely reduces the effectiveness of the model for data augmentation, as the synthesized data fails to reflect the appearance diversity and visual quality found in diverse real-world environments, such as varying weather conditions, lighting, water motion, or terrain texture. 

To address this issue, we aim to enhance the diversity of the generated distribution:
\begin{equation}
    p_{\text{div}}(\mathbf{x} \mid \mathbf{M}, \mathbf{t}_{\text{style}})
\end{equation}
by introducing two components at inference time:
\begin{itemize}
    \item \textbf{Class-Aware Style Bank} $\mathbf{t}_{\text{style}}$: injects structured appearance variation relevant to each semantic class.
    \item \textbf{Adaptive Annealing Sampling (AAS)}: modulates the influence of the conditioning signal throughout the sampling process to promote mode diversity without sacrificing semantic alignment.
\end{itemize}

\noindent The proposed method can be applied to any conditional diffusion framework that supports spatial control, such as ControlNet++~\cite{li2024controlnet++}, Uni-ControlNet~\cite{zhao2023uni}, and T2I-Adapter~\cite{mou2024t2i}. We focus on inference-time mechanisms, so no retraining of these backbones is required.

\subsection{Class-Aware Style Bank}
The Style Bank is a key component of our inference-time framework that introduces appearance-level diversity into the conditional diffusion process while preserving structural constraints defined by the input segmentation mask. In the domain-specific task of marine obstacle detection, the number of semantic classes is limited, and textual descriptions tend to be repetitive (e.g., ``sky'', ``water'', ``obstacle''). This limits the effectiveness of prompt-based diversity. To address this, we construct an interpretable style bank that maps each class to a set of stylistic descriptors reflecting real-world visual variations.

The style bank $\mathcal{S}$ consists of:
\begin{itemize}
    \item \textit{Class-specific style descriptors} $\mathcal{S}_c = \{s_{c,1}, s_{c,2}, ..., s_{c,k}\}$ for $c \in \mathcal{C}$.
    \item \textit{Global scene descriptors} $\mathcal{G}$ that represent lighting/atmosphere: $\mathcal{G} = \{g_1, g_2, ..., g_m\}$.
\end{itemize}

\begin{algorithm}[t]
\caption{Style-Enhanced Prompt Construction}
\label{alg:prompt_construction}
\begin{algorithmic}[1]
\Require
    Segmentation mask $\mathbf{M}$ \\
    Ordered class list: $\mathcal{C} = \{c_1, c_2, \dots, c_k\}$ extracted from $\mathbf{M}$ \\
    $\mathcal{S}$: Style bank $\{ \mathcal{S}_{c_1}, \mathcal{S}_{c_2}, \dots, \mathcal{S}_{c_k} \}$ \\
    $\mathcal{G}$: Global descriptor set $\{g_1, g_2, ..., g_m\}$
\Ensure
    Stylized prompt $\mathbf{t}_{\text{style}}$
\State Identify present classes: $\mathcal{P} \gets \{c \in \mathcal{C} \mid c ~\text{presents in}~\mathbf{M}\}$
\State Initialize prompt string: $\mathbf{t} \gets \textit{``This image contains''}$
\State Initialize style phrase list: $\mathcal{O} \gets [~]$
\For{each $c_i \in \mathcal{C}$ \textbf{in specified order}}
    \If{$c_i \in \mathcal{P}$}
        \State Sample descriptor: $s_i \gets$  \text{random element from } $\mathcal{S}_{c_i}$
        \State Construct phrase: $\phi \gets s_i + \text{`` "} + c_i$
        \State Append to list: $\mathcal{O} \gets \mathcal{O} \cup \{\phi\}$
    \EndIf
\EndFor

\If{$|\mathcal{O}| = 1$}
    \State $\mathbf{t} \gets \mathbf{t} + \mathcal{O}[0]$
\Else
    \State $\mathbf{t} \gets \mathbf{t} + \text{join}(\mathcal{O}[0\!:\!-1], \text{``, "})$
    \State $\mathbf{t} \gets \mathbf{t} + \text{`` and "} + \mathcal{O}[-1]$
\EndIf
\State Sample global descriptor: $g \gets \text{random choice from } \mathcal{G}$
\State Finalize prompt: $\mathbf{t} \gets \mathbf{t} + \text{``, "} + g + \text{``."}$\\
\Return $\mathbf{t}_{\text{style}} \gets \mathbf{t}$
\end{algorithmic}
\end{algorithm}

\noindent Given a set of classes $\mathcal{C} = \{c_1, c_2, \dots, c_k\}$  present in the input segmentation mask $\mathbf{M}$, we sample a descriptor $s_i \in \mathcal{S}_{c_i}$ for each class, and sample a global descriptor $g \in \mathcal{G}$. The resulting style-enhanced prompt is constructed by combining these class-wise phrases with the global scene descriptor summarized in Algorithm~\ref{alg:prompt_construction}.

In our current implementation, sampling from both $\mathcal{S}$ and $\mathcal{G}$ is performed uniformly. However, the proposed framework is designed to support more advanced strategies in future work, such as learned, user-guided sampling. This flexibility enables controllable and semantically aligned diversity in the generated images. 

While our current study focuses on marine obstacle detection, the proposed framework is task-agnostic and applicable to other conditional generation problems. Any task that benefits from class-conditioned diversity and structured textual prompts, such as urban scene understanding, agricultural monitoring, or medical image synthesis, can leverage the same style-driven augmentation mechanism by defining appropriate class-wise and global style descriptors.

\subsection{Adaptive Annealing Sampling}
While the style bank enhances semantic diversity through prompt engineering, we further improve visual diversity during image generation through adaptive conditioning perturbation and noise scheduling. Standard diffusion sampling with fixed prompts often produces visually similar outputs, especially with structurally constrained models such as ControlNet~\cite{zhang2023adding}. To address this, we propose an Adaptive Annealing Sampling strategy that enhances diversity through structural conditioning annealing.


Given text embeddings $\mathbf{y}_p$ (positive) and $\mathbf{y}_u$ (unconditional), we inject anti-correlated noise with adaptive scale $s(t)$ and residual jitter $\sigma_r$ to encourage early exploration while preserving semantic coherence.

\begin{align}
\tilde{\mathbf{y}}_p &= \sqrt{\gamma(t)}\mathbf{y}_p - s(t)\sqrt{1-\gamma(t)}\mathbf{n} + \boldsymbol{\epsilon}_p, \quad \boldsymbol{\epsilon}_p \sim \mathcal{N}(0,\sigma_r^2\mathbf{I}) \\
\tilde{\mathbf{y}}_u &= \sqrt{\gamma(t)}\mathbf{y}_u + s(t)\sqrt{1-\gamma(t)}\mathbf{n} + \boldsymbol{\epsilon}_u, \quad \boldsymbol{\epsilon}_u \sim \mathcal{N}(0,\sigma_r^2\mathbf{I})
\end{align}

\noindent\textbf{Adaptive Control Mechanism:} At each step $t$:
\begin{enumerate}
    \item Compute \textit{Conditional Output Discrepancy (COD)} via dual-path perturbation:
    \begin{align}
        \text{COD}(t) = \| D(\mathbf{z}_t, \mathbf{M}, \tilde{\mathbf{y}}_p^1) - D(\mathbf{z}_t, \mathbf{M}, \tilde{\mathbf{y}}_p^2) \|_2
    \end{align}
    where $\tilde{\mathbf{y}}_p^{1,2}$ are independently perturbed positive embeddings
    \item Update noise scale with proportional-integral control:
    \begin{equation}
        s(t+1) = \text{clip}\left( s(t) + \kappa_p \left( \tau - \text{COD}(t) \right), s_{\min}, s_{\max} \right)
    \end{equation}
    \item Anneal ControlNet conditioning strength:
    \begin{equation}
        \lambda(t) = \lambda_{\min} + (\lambda_{\max} - \lambda_{\min}) \gamma(t)
    \end{equation}
\end{enumerate}

\begin{algorithm}[t]
\caption{Adaptive Annealing Sampling (AAS).}
\label{alg:aas}
\begin{algorithmic}[1]
\Require \\
    $\mathbf{z}_T$ : Initial latent \\
    $\mathbf{y}_p$ : Positive text embedding \\
    $\mathbf{y}_u$ : Unconditional text embedding \\
    $\mathbf{M}$ : Segmentation mask \\
    $\tau_1, \tau_2$ : Annealing bounds ($0 \leq \tau_1 < \tau_2 \leq T$) \\
    $\tau = 0.1$ : Target COD \\
    $\kappa_p = 0.01$ : Proportional gain \\
    $s_{\text{init}} = 0.1$ : Initial noise scale \\
    $s_{\min} = 0.05, s_{\max} = 0.5$ : Noise bounds \\
    $\sigma_r = 0.1$ : Residual noise \\
    $\lambda_{\min} = 0.6, \lambda_{\max} = 1.0$ : ControlNet bounds

\Ensure $\mathbf{z}_0$ : Generated sample
\State $s \gets s_{\text{init}}$
\For{$t = T \to 1$}
    \State $\gamma \gets \textsc{LinearSchedule}(t, \tau_1, \tau_2)$ \Comment{$\gamma:1.0\to0.0$}
    \State $\lambda \gets \lambda_{\min} + (\lambda_{\max} - \lambda_{\min}) \gamma$
    
    \State // \textbf{COD estimation}
    \State $\tilde{\mathbf{y}}_p^1 \gets \sqrt{\gamma}\mathbf{y}_p + s\sqrt{1-\gamma}\mathbf{n}_1$
    \State $\tilde{\mathbf{y}}_p^2 \gets \sqrt{\gamma}\mathbf{y}_p + s\sqrt{1-\gamma}\mathbf{n}_2$, $\mathbf{n}_1, \mathbf{n}_2 \sim \mathcal{N}(0, \mathbf{I})$
    \State $\hat{\epsilon}_1 \gets D(\mathbf{z}_t, t, \mathbf{M}, \tilde{\mathbf{y}}_p^1; \lambda)$
    \State $\hat{\epsilon}_2 \gets D(\mathbf{z}_t, t, \mathbf{M}, \tilde{\mathbf{y}}_p^2; \lambda)$
    \State $\text{COD} \gets \|\hat{\epsilon}_1 - \hat{\epsilon}_2\|_2$
    
    \State // \textbf{Noise scale update}
    \State $s' \gets \textsc{Clip}(s + \kappa_p (\tau - \text{COD}), s_{\min}, s_{\max})$
    
    \State \textbf{// Anti-correlated perturbation}
    \State $\tilde{\mathbf{y}}_p \gets \sqrt{\gamma}\mathbf{y}_p - s\sqrt{1-\gamma}\mathbf{n} + \sigma_r\boldsymbol{\epsilon}_p$, 
    \State $\tilde{\mathbf{y}}_u \gets \sqrt{\gamma}\mathbf{y}_u + s\sqrt{1-\gamma}\mathbf{n} + \sigma_r\boldsymbol{\epsilon}_u$, $\mathbf{n} \sim \mathcal{N}(0, \mathbf{I})$
    
    \State // \textbf{Classifier-free guidance}
    \State $\hat{\epsilon}_u \gets D(\mathbf{z}_t, t, \mathbf{M}, \tilde{\mathbf{y}}_u; \lambda)$
    \State $\hat{\epsilon}_p \gets D(\mathbf{z}_t, t, \mathbf{M}, \tilde{\mathbf{y}}_p; \lambda)$
    \State $\hat{\epsilon} \gets \hat{\epsilon}_u + w \cdot (\hat{\epsilon}_p - \hat{\epsilon}_u)$ \Comment{Guidance scale $w$}
    
    \State \textbf{Scheduler step}~~$\mathbf{z}_{t-1} \gets \textsc{Scheduler:} (\mathbf{z}_t, \hat{\epsilon}, t)$
    \State $s \gets s'$ \Comment{Update for next step}
\EndFor\\
\Return $\mathbf{z}_0$
\end{algorithmic}
\end{algorithm}

\noindent\textbf{Key Implementation Details:}
\begin{itemize}
    \item \textbf{Anti-correlated noise}: Maintains semantic coherence while exploring diverse outputs (Eqs. 1 and 2).
    \item \textbf{Residual noise} ($\sigma_r = 0.1$): Prevents mode collapse (Algorithm line 20).
    \item \textbf{ControlNet annealing}: $\lambda(t)$ from 0.6 to 1.0 maintains structural fidelity (Eq. 5).
    \item \textbf{Linear schedule}: $\gamma(t) = \max\left(0, \min\left(1, \frac{\tau_2 - t}{\tau_2 - \tau_1}\right)\right)$
    \item \textbf{Adaptive bounds}: $s \in [0.05, 0.5]$ ensures numerical stability
\end{itemize}

\noindent\textbf{Theoretical Convergence}: As $\gamma(t) \to 1$ during late denoising:
\begin{itemize}
    \item $\text{COD}(t) \to 0$ by continuity of $D(\cdot)$
    \item $\lambda(t) \to 1$ ensures strict adherence to structural constraints
    \item $s(t) \to s_{\min}$ minimizes stochastic variation
\end{itemize}

This guarantees convergence to mask-consistent outputs while early denoising steps explore diverse appearances.

The proposed adaptive annealing sampling strategy is inspired by the CADS framework~\cite{sadat2023cads} but improves it in three main aspects: 
(1) Unlike the CADS's static noise schedule, our closed-loop control dynamically adjusts noise scales via Conditional Output Discrepancy (COD) feedback, enabling sample-specific diversity-fidelity balancing; 
(2) We introduce structural conditioning annealing with $\lambda(t) = 0.6 + 0.4\gamma(t)$ to progressively relax ControlNet constraints, maintaining mask consistency while exploring diverse appearances; 
and (3) Anti-correlated perturbation with residual noise $\sigma_r=0.1$ prevents mode collapse without compromising semantic coherence. The complete Adaptive Annealing Sampling strategy is summarized in Algorithm \ref{alg:aas}.

\section{Experiments}
We evaluate our proposed framework on one real-world, domain-specific dataset about marine obstacle detection with three core semantic categories: obstacle, sky, and water. This task poses challenges for generative modeling due to limited linguistic variability and the need for semantically faithful, visually realistic, and diverse training samples. During training, we use simple class-list prompts to ensure clean conditional alignment. At inference, we apply a dataset-specific style bank combined with the proposed AAS to generate high-quality, layout-consistent synthetic images. We evaluate the effectiveness of our method in terms of its impact on downstream segmentation performance when using generated images for data augmentation.

\subsection{Datasets}
\noindent\textbf{MaSTr1325:} Marine Semantic Segmentation Training Dataset: MaSTr1325 is a new large-scale marine semantic segmentation training dataset tailored for the development of obstacle detection methods in small-sized coastal USVs. The dataset contains 1325 diverse images captured over a two-year span with a real USV, covering a range of realistic conditions encountered in a coastal surveillance task. All images are per-pixel semantically labeled and synchronized with inertial measurements of the on-board sensors. In addition, a dataset augmentation protocol is proposed to address slight appearance differences of the images in the training set and those in deployment.

\noindent\textbf{MODS:} The MODS~\cite{bovcon2021mods} dataset is a comprehensive benchmark for unmanned surface vehicle (USV) perception. It unifies three prior datasets: MODD1~\cite{kristan2015fast}, MODD2~\cite{bovcon2018stereo}, and SMD~\cite{prasad2017video}, and comprises 94 sequences, and over 80k stereo images, with 63k annotated obstacles and 10k water-edge annotations, captured across diverse real-world coastal scenarios. MODS provides standardized evaluation protocols for both object detection and obstacle segmentation.

\subsection{Experimental Setup}
Our framework builds on ControlNet++~\cite{li2024controlnet++}, a strong baseline for controllable image generation with pixel-level condition consistency. We fine-tune ControlNet++ using segmentation masks as input conditions and simplified class-list text prompts, such as ``this image contains sky, water, obstacle,'' which reflect the fixed and limited class space of our target datasets. We retain the ControlNet++ architecture and training procedure, but introduce no additional diversity mechanisms during training to maintain strict semantic alignment. During inference, we incorporate our proposed dataset-specific style bank and apply Adaptive Annealing Sampling (AAS) to enhance output diversity. The style bank contains domain-relevant style phrases per class, and is adapted to each dataset. 

All models are trained on the MaSTr1325 dataset, which contains high-quality pixel-level semantic segmentation annotations. For evaluation, we use the MODS dataset, which lacks pixel-wise ground truth masks but provides water-edge annotations and bounding boxes for dynamic obstacles. As a result, we evaluate model performance using proxy detection-based metrics such as water-edge RMSE, water-land detection rate, precision, recall, true/false positives, and F1 score~\cite{bovcon2021mods}. This setup allows us to assess real-world generalization in a weakly annotated test domain, and evaluate how the visual quality and diversity of generative training samples affect segmentation performance. We note that our proposed Style Bank and AAS components add no extra retraining beyond this baseline fine-tuning.

\subsection{Inference Strategy Design}

To fairly evaluate the effectiveness of our proposed method, we design a controlled experiment built upon the ControlNet++ framework, isolating the contributions of the Style Bank prompt mechanism and the Adaptive Annealing Sampling (AAS) strategy. The ControlNet++ model is first trained on our custom dataset using a fixed training configuration, with basic conditioning prompts such as ``This image contains sky, water, and obstacle.'' No prompt diversity or style guidance is introduced during the training phase.

At the inference stage, we generate synthetic data using the following three strategies:

\begin{itemize}
    \item \textbf{Traditional Image Augmentation (Traditional):} Label-preserving augmentations are applied directly to the original training data, generating four variants per image using standard techniques including horizontal flip, small-angle rotation (±5°), brightness and contrast adjustment, affine shifting and scaling, and mild blur or noise. These augmentations operate at the image level and do not introduce semantic or stylistic variation.
    
    \item \textbf{Default Generation (Default Gen):} ControlNet++ is used with its original inference setting, relying on simple prompts and standard deterministic sampling (e.g., DDIM). This setting lacks stylistic control or adaptive sampling dynamics.
    
    \item \textbf{Our Inference-Time Strategy (Our Gen):} We extend ControlNet++ by introducing class-aware prompts from a carefully selected Style Bank that reflect diverse maritime conditions (e.g., ``a calm water surface under a stormy sky with distant obstacles''). We further apply AAS, which dynamically adjusts the noise scale and conditioning strength during generation to balance visual diversity and layout consistency. This method enhances the visual quality, diversity, and task relevance of generated samples entirely at inference time, without retraining the model.
\end{itemize}

By evaluating all three strategies under a downstream testing pipeline, we ensure a fair comparison. This setup highlights the effectiveness of our approach in generating semantically controlled, layout-consistent training samples that improve downstream performance.

\definecolor{mygray}{gray}{0.9}
\begin{table*}[htbp]
\centering
\caption{Evaluation results across two datasets and different augmentation settings. Reported metrics include water-edge RMSE (px), water-land detections (\%), precision (Pr), recall (Re), true positives (TPr), false positives (FPr), and F1 score.}
\label{tab:our_benchmark}
\scalebox{0.75}{
\begin{tabular}{llccccccc}
\toprule
\textbf{Architecture} & \textbf{Setting} 
& \textbf{Water-edge RMSE$\downarrow$} 
& \textbf{Water-Land det. (\%)$\uparrow$} 
& \textbf{Pr (\%)$\uparrow$} 
& \textbf{Re (\%)$\uparrow$} 
& \textbf{TPr$\uparrow$} 
& \textbf{FPr$\downarrow$} 
& \textbf{F1 (\%)$\uparrow$} \\
\midrule

\multirow{12}{*}{DeepLabV3+} 
& Original (kope)     & 37 & 94.5 & 87.5 & 98.8 & 29.1 & 8.1 & 92.8 \\
& Original (stu)     & 16 & 98.3 & 87.9 & 96.4 & 42.2 & 22.8 & 92.0 \\
& Original (All)          & 22 & 97.3 & 87.8 & 97.2 & 38.8 & 18.9 & 92.2 \\
\rowcolor{mygray}
& + Traditional (kope) & 19 & 97.9 & 89.5 & 96.0 & 41.9 & 21.1 & 92.6 \\
\rowcolor{mygray}
& + Traditional (stu) & 46 & 92.8 & 88.1 & 98.2 & 29.1 & 9.7 & 92.9 \\
& \cellcolor{mygray}+ Traditional (All)      & \cellcolor{mygray}26 & \cellcolor{mygray}96.6 & \cellcolor{mygray}89.0 & \cellcolor{mygray}96.7 & \cellcolor{mygray}38.5 & \cellcolor{mygray}18.1 & \cellcolor{mygray}92.7 \\
& + Default Gen (kope) & 22 & 97.3 & 88.5 & 93.5 & 41.2 & 12.7 & 90.9 \\
& + Default Gen (stu) & 46 & 93.2 & 89.5 & 96.5 & 28.5 & 3.8 & 92.9 \\
& + Default Gen (All)      & 28 & 96.2 & 88.8 & 94.4 & 37.9 & 10.4 & 91.5 \\
\rowcolor{mygray}
& + Our Gen (kope)     & 17 & 98.2 & 94.5 & 93.8 & 41.2 & 8.5 & 94.2 \\
\rowcolor{mygray}
& + Our Gen (stu)     & 44 & 93.9 & 90.4 & 97.6 & 28.8 & 7.4 & 93.9 \\
\rowcolor{mygray}
& + Our Gen (All)          & 26 & 96.6 & 93.1 & 95.0 & 38.0 & 8.2 & 92.1 \\
\midrule

\multirow{12}{*}{Mask2Former} 
& Original (kope)     & 82 & 87.4 & 73.2 & 98.0 & 42.8 & 35.8 & 83.8 \\
& Original (stu)     & 85 & 86.2 & 81.0 & 98.4 & 29.2 & 11.1 & 88.9 \\
& Original (All)          & 83 & 87.1 & 75.5 & 98.1 & 39.2 & 29.3 & 85.3 \\
\rowcolor{mygray}
& + Traditional (kope) & 84 & 86.9 & 78.8 & 95.8 & 42.2 & 32.7 & 86.5 \\
\rowcolor{mygray}
& + Traditional (stu) & 99 & 83.2 & 84.8 & 92.4 & 29.2 & 10.2 & 88.4 \\
& \cellcolor{mygray}+ Traditional (All)      & \cellcolor{mygray}88 & \cellcolor{mygray}85.9 & \cellcolor{mygray}80.5 & \cellcolor{mygray}94.7 & \cellcolor{mygray}38.8 & \cellcolor{mygray}26.8 & \cellcolor{mygray}87.0 \\
& + Default Gen (kope) & 148 & 74.2 & 76.8 & 93.1 & 41.3 & 37.8 & 84.4 \\
& + Default Gen (stu) & 132 & 74.4 & 81.9 & 86.6 & 29.2 & 13.7 & 84.2 \\
& + Default Gen (All)      & 144 & 74.3 & 78.6 & 91.1 & 38.2 & 29.7 & 84.4 \\
\rowcolor{mygray}
& + Our Gen (kope)     & 105 & 84.6 & 77.2 & 95.8 & 42.0 & 34.0 & 85.2 \\
\rowcolor{mygray}
& + Our Gen (stu)     & 139 & 75.0 & 88.0 & 89.3 & 29.2 & 94.2 & 88.7 \\
\rowcolor{mygray}
& + Our Gen (All)          & 114 & 82.1 & 79.8 & 93.7 & 38.6 & 28.7 & 86.2 \\
\midrule

\multirow{12}{*}{SAM-Adapter} 
& Original (kope)     & 13 & 98.7 & 93.7 & 96.3 & 42.8 & 4.8 & 95.0 \\
& Original (stu)     & 23 & 97.4 & 90.9 & 94.8 & 28.6 & 3.1 & 92.8 \\
& Original (All)          & 16 & 98.3 & 92.8 & 95.8 & 39.1 & 4.4 & 94.3 \\
\rowcolor{mygray}
& + Traditional (kope) & 16 & 97.9 & 95.1 & 97.1 & 43.1 & 6.6 & 96.1 \\
\rowcolor{mygray}
& + Traditional (stu) & 17 & 98.4 & 93.9 & 95.5 & 28.9 & 3.1 & 94.7 \\

& \cellcolor{mygray}+ Traditional (All)      & \cellcolor{mygray}16 & \cellcolor{mygray}98.0 & \cellcolor{mygray}94.7 & \cellcolor{mygray}96.6 & \cellcolor{mygray}39.4 & \cellcolor{mygray}5.7 & \cellcolor{mygray}95.7 \\
& + Default Gen (kope) & 18 & 97.7 & 96.6 & 96.2 & 42.1 & 2.6 & 96.4 \\
& + Default Gen (stu) & 19 & 98.1 & 89.2 & 98.4 & 28.4 & 3.3 & 93.6 \\
& + Default Gen (All)      & 18 & 97.8 & 94.1 & 96.9 & 38.5 & 2.8 & 95.5 \\
\rowcolor{mygray}
& + Our Gen (kope)     & 13 & 98.7 & 95.1 & 96.7 & 42.7 & 7.6 & 95.9 \\
\rowcolor{mygray}
& + Our Gen (stu)     & 17 & 97.9 & 93.0 & 96.9 & 29.0 & 5.3 & 94.9 \\
\rowcolor{mygray}
& + Our Gen (All)          & 14 & 98.5 & 94.4 & 96.7 & 39.1 & 7.0 & 95.6 \\
\bottomrule
\end{tabular}
}
\end{table*}

\subsection{Quantitative Results}

Table~\ref{tab:our_benchmark} presents comparisons of three segmentation models--DeepLabV3+~\cite{mmseg2020}, Mask2Former~\cite{mmseg2020}, and SAM-Adapter~\cite{chen2023sam}--trained on different data augmentation settings and evaluated on two marine obstacle detection datasets: kope and stu in the MODS dataset. The “All” setting denotes evaluation on the combined dataset of kope and stu.

We compare four augmentation strategies: 
(1) \textbf{Original}, where no augmentation is applied; 
(2) \textbf{Traditional}, using standard geometric and photometric transformations; 
(3) \textbf{Default Gen}, where samples are generated using ControlNet++ with simple prompts; and 
(4) \textbf{Our Gen}, the proposed method combining class-aware Style Bank prompts and Adaptive Annealing Sampling (AAS). All models are trained from scratch on the augmented datasets accordingly, using identical training protocols.

Across all architectures and datasets, our method consistently delivers strong performance, particularly in terms of precision, recall, and F1 score, while maintaining low water-edge RMSE and reduced false positives. For instance, on DeepLabV3+ trained with the combined dataset (“All”), Our Gen achieves an F1 score of 92.1\%, which is competitive with Traditional augmentation (92.7\%) and better than Default Gen (91.5\%). Notably, it yields higher precision (93.1\%) and lower false positive rate (8.2), reflecting the effectiveness of the generated samples.

SAM-Adapter achieves the highest F1 scores across all settings. However, our augmentation method offers consistent benefits over other augmentation strategies. For example, with SAM-Adapter on the combined dataset, our Gen achieves an F1 score of 95.6\%, outperforming Default Gen (95.5\%) and matching Traditional (95.7\%), while offering superior recall (96.7\%) and a lower false positive rate than the Original setting. This illustrates that our method provides a favorable trade-off between sensitivity and precision, which is crucial for reliable obstacle detection.

\begin{figure*}[t]
  \centering
   \includegraphics[width=0.8\linewidth]{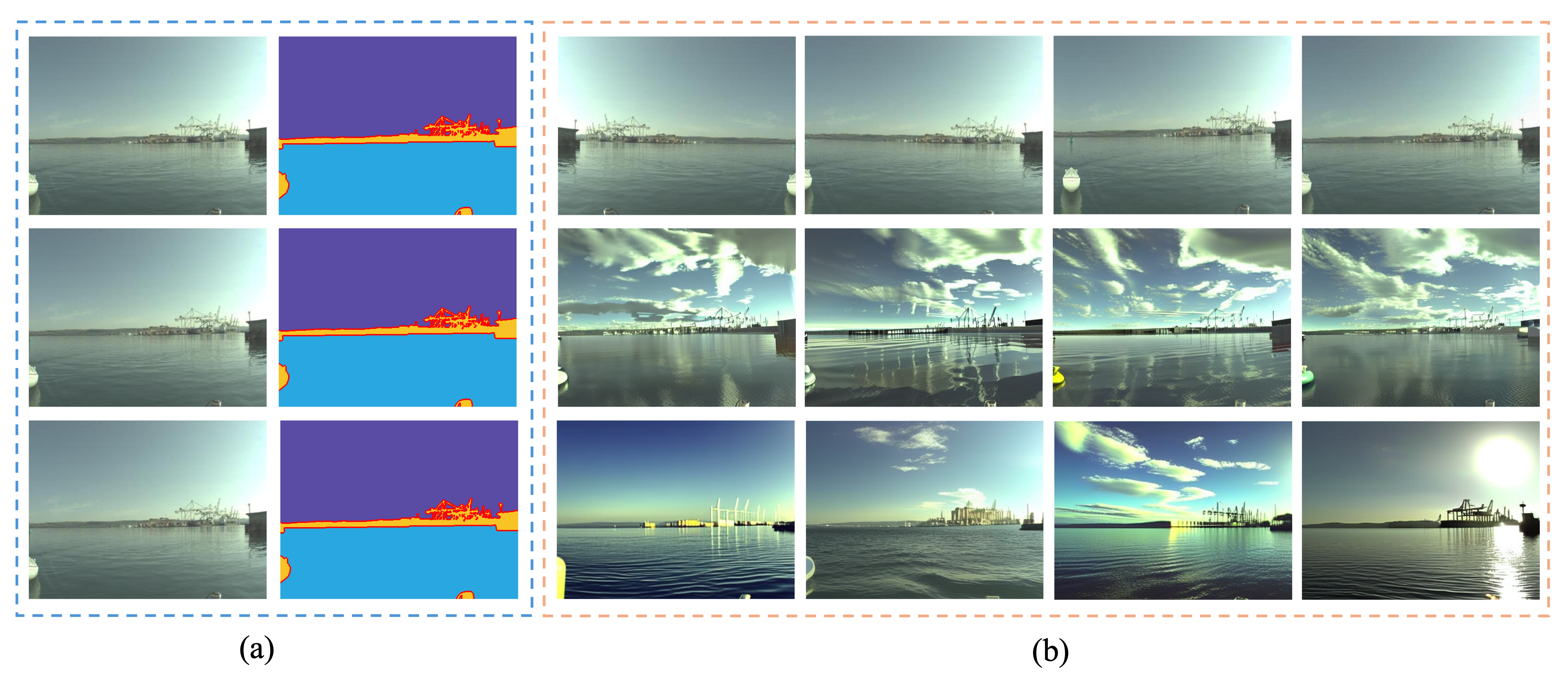}
    \vspace{-0.5cm}
   \caption{Example of augmented images. 
   (a) Original example image and mask; 
   (b) Augmented images. 1st row: generated images with traditional data augmentation method; 2nd row: Generated images using Default setting of ControlNet ++(simple prompt); 3rd row: Generated images using proposed AAS method.}
   \label{fig:sample1}
\end{figure*}
\begin{figure*}[t]
  \centering
   \includegraphics[width=0.8\linewidth]{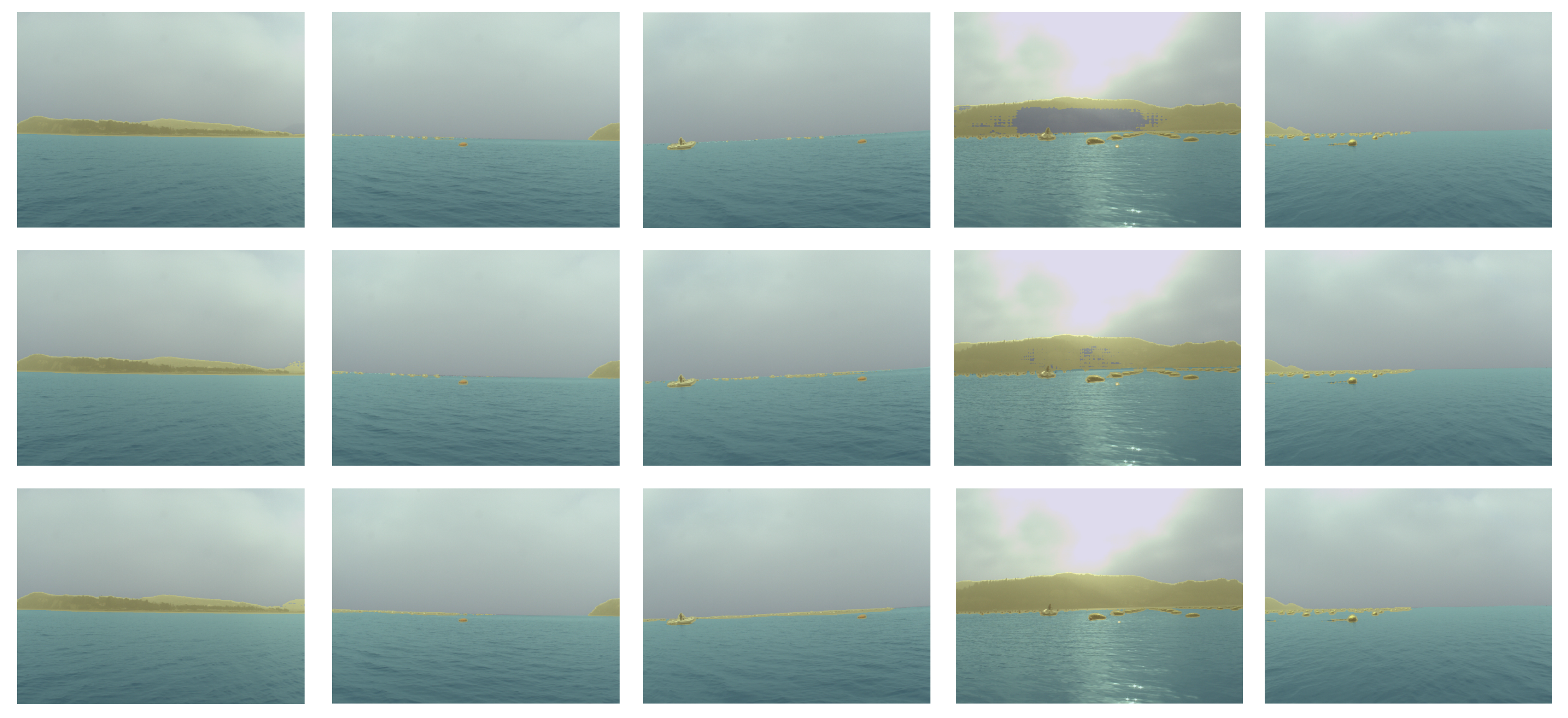}
   \caption{Overlayed images from traditional method (1st row), simple prompt (2nd row), and the proposed AAS method (3rd row).
   }
   \label{fig:sample2}
\end{figure*}

For Mask2Former, known for higher RMSE and more volatile detection behavior, Our Gen improves performance over Default Gen in most metrics. On the combined dataset, Our Gen yields an F1 score of 86.2\%, outperforming Default Gen (84.4\%) and approaching the Traditional baseline (87.0\%), while also improving precision and recall score.

Importantly, the proposed method operates entirely at inference time during data generation and improves both the perceptual quality and semantic diversity of training samples. By injecting class-aware appearance variation and modulating sampling to maintain mask fidelity, it achieves robust results on complex downstream segmentation tasks. These results highlight the strength of our augmentation pipeline in generating layout-consistent, class-aware samples that enhance segmentation model performance.

\subsection{Qualitative results}
\vspace{-0.1cm}
Figure~\ref{fig:sample1} illustrates synthetic images generated by three augmentation strategies. While traditional augmentation introduces limited pixel-level variation, it does not alter scene semantics or layout, and lacks the capacity to generate visually realistic, context-aware training samples. 
Default Generation introduces some appearance variation but can lack semantic fidelity and may produce artifacts. In contrast, our method produces high-quality, visually diverse, semantically meaningful, and layout-consistent samples that better reflect real-world maritime scenes.
Figure~\ref{fig:sample2} presents segmentation outputs predicted by models trained on these augmented datasets. Traditional augmentation frequently misses large obstacles, while Default Generation struggles with completeness and false positives. Our method enables more accurate and complete obstacle detection, even under challenging conditions such as fog, demonstrating the importance of semantically rich and stylistically realistic training data for robust segmentation performance.
\vspace{-0.2cm}
\section{Conclusions}
We proposed an inference-time data augmentation strategy, combining class-aware Style Bank prompts with Adaptive Annealing Sampling (AAS). Our method introduces visual quality, semantic diversity, and layout fidelity into the generated training data without requiring retraining of the generative model. Experiments on two marine obstacle detection datasets using three segmentation models show that our approach achieves competitive or improved performance across different metrics. Compared to traditional and default generation methods, our strategy yields more consistent and layout-aware synthetic data, enhancing downstream segmentation.